%% file: nips2015_transfer.tex
\documentclass{article} % For LaTeX2e
\usepackage{nips15submit_e,times}
\usepackage{hyperref}
\usepackage{url}
\usepackage{color}
\usepackage{amsmath,amssymb} % define this before the line numbering.
\usepackage{graphicx}
\usepackage{enumitem}
\usepackage{tabularx}
\usepackage{arydshln}
\newcolumntype{C}[1]{>{\centering\let\newline\\\arraybackslash\hspace{0pt}}p{#1}}

\newcommand{\x}{\mathbf{x}}
\newcommand{\y}{\mathbf{y}}
\newcommand{\z}{\mathbf{z}}
\newcommand{\g}{\mathbf{g}}
\newcommand{\f}{\mathbf{f}}

\title{Decoupled Deep Neural Network for \\Semi-supervised Semantic Segmentation}
\author{
Seunghoon Hong, \ Hyeonwoo Noh, \ Bohyung Han \\
Dept. of Computer Science and Engineering, POSTECH, Pohang, Korea \\
\texttt{\{maga33,hyeonwoonoh\_,bhhan\}@postech.ac.kr}\\
}

\nipsfinalcopy % Uncomment for camera-ready version

\begin{document}
\maketitle

% ====================================================================
% Abstract
\input{abstract.tex}

% ====================================================================
% Introduction
\input{introduction.tex}

% ====================================================================
% Related Work
\input{relatedwork.tex}

% ====================================================================
% Algorithm overview
\input{overview.tex}

% ====================================================================
% System Architecture
%\input{architecture.tex}
%\input{architecture_clsbrgseg.tex}
\input{architecture_clssegbrg.tex}

% ====================================================================
% Tranfser Learning
%\input{transfer.tex}

% ====================================================================
% Training and Inference using the network
\input{trainAndInference.tex}

% ====================================================================
% Experiments
\input{experiments.tex}

% ====================================================================
% Conclusion
\input{conclusion.tex}

{\small
\bibliographystyle{unsrt}
\bibliography{egbib}
}

\end{document}

%% file: abstract.tex
% !TEX root = nips2015_transfer.tex
\begin{abstract}
We propose a novel deep neural network architecture for semi-supervised semantic segmentation using heterogeneous annotations.
Contrary to existing approaches posing semantic segmentation as a single task of region-based classification, our algorithm decouples classification and segmentation, and learns a separate network for each task.
%Contrary to existing approaches posing semantic segmentation by region-based classification, the proposed architecture decouples classification and segmentation by two separate networks.
%The proposed architecture is composed of two separate networks for classification and segmentation. 
%Posing semantic segmentation as separate problems of classification and segmentation, we develop a novel deep architecture composed of separate networks for classification and segmentation.
%Posing semantic segmentation as separate problems of classification and segmentation, our network achieves the two objectives by separate networks for classification and segmentation.
%Our algorithm achieves semantic segmentation by separate networks for classification and segmentation.
In this architecture, labels associated with an image are identified by classification network, and binary segmentation is subsequently performed for each identified label in segmentation network.
%With this model, classification network identifies labels associated with an input image, and segmentation network produces binary segmentation corresponding to each identified label.
%This decoupling is appropriate to exploit heterogeneous annotations of semi-supervised setting, where image-level and pixel-wise class labels are used to train the classification and segmentation network, respectively.
The decoupled architecture enables us to learn classification and segmentation networks separately based on the training data with image-level and pixel-wise class labels, respectively.
It facilitates to reduce search space for segmentation effectively by exploiting class-specific activation maps obtained from bridging layers.
Our algorithm shows outstanding performance compared to other semi-supervised approaches even with much less training images with strong annotations in PASCAL VOC dataset.
%Our network demonstrates outstanding performance in PASCAL VOC dataset, which is substantially better than existing semi-supervised approaches with much smaller examples with pixel-wise annotations. 
%We pose the semantic segmentation as separate problems of classification and segmentation, and develop a novel architecture composed of separate networks for classification and segmentation; 
%in this formulation, semantic segmentation is performed by independent and successive operations of classification and segmentation, where labels associated with an image are identified by classification network, and segmentation mask corresponding to each identified label is obtained by segmentation network.
%Given these networks, semantic segmentation is performed by identifying associated labels of input image by classification network, and subsequently construct segmentation mask of each identified label by segmentation network.
%The decoupling of classification and segmentation is appropriate to train each network using heterogeneous annotations given by image-level and pixel-level class labels 
\end{abstract}

%% file: introduction.tex
	% !TEX root = nips2015_transfer.tex
\section{Introduction}

\ifdefined\paratitle {\color{blue} 
[Introductory descriptions of semantic segmentation problem.]\\
} \fi
Semantic segmentation is a technique to assign structured semantic labels---typically, object class labels---to individual pixels in images. 
This problem has been studied extensively over decades, yet remains challenging since object appearances involve significant variations that are potentially originated from pose variations, scale changes, occlusion, background clutter, etc.
However, in spite of such challenges, the techniques based on Deep Neural Network (DNN) demonstrate impressive performance in the standard benchmark datasets such as PASCAL VOC~\cite{Pascalvoc}.
%However, in spite of such challenges, the techniques based on Deep Neural Network (DNN) accelerate the advance of semantic segmentation algorithms and demonstrate impressive performance in the standard benchmark datasets such as PASCAL VOC~\cite{Pascalvoc}.

\ifdefined\paratitle {\color{blue} 
[Supervised DNNs for semantic segmentation]\\
} \fi
Most DNN-based approaches pose semantic segmentation as pixel-wise classification problem~\cite{Fcn,Deeplabcrf,Hypercolumns,Sds,Zoomout}. 
Although these approaches have achieved good performance compared to previous methods, training DNN requires a large number of segmentation ground-truths, which result from tremendous annotation efforts and costs. 
%the performance of DNN heavily depends on size and quality of annotations.
For this reason, reliable pixel-wise segmentation annotations are typically available only for a small number of classes and images, which makes supervised DNNs difficult to be applied to semantic segmentation tasks involving various kinds of objects.

\ifdefined\paratitle {\color{blue} 
[Semi and weakly supervised DNNs for semantic segmentation]\\
} \fi
Semi- or weakly-supervised learning approaches~\cite{Wsl,Wssl,Fcmil,Boxsup} alleviate the problem in lack of training data by exploiting weak label annotations per bounding box~\cite{Boxsup,Wssl} or image~\cite{Wsl,Wssl,Fcmil}. 
They often assume that a large set of weak annotations is available during training while training examples with strong annotations are missing or limited.
This is a reasonable assumption because weak annotations such as class labels for bounding boxes and images require only a fraction of efforts compared to strong annotations, {\it i.e.,} pixel-wise segmentations.
The standard approach in this setting is to update the model of a supervised DNN by iteratively inferring and refining hypothetical segmentation labels using weakly annotated images.
Such iterative techniques often work well in practice~\cite{Wssl,Boxsup}, but training methods rely on ad-hoc procedures and there is no guarantee of convergence; implementation may be tricky and the algorithm may not be straightforward to reproduce.

\ifdefined\paratitle {\color{blue} 
[Our approach]\\
} \fi
We propose a novel decoupled architecture of DNN appropriate for semi-supervised semantic segmentation, which exploits heterogeneous annotations with a small number of strong annotations---full segmentation masks---as well as a large number of weak annotations---object class labels per image.
Our algorithm stands out from the traditional DNN-based techniques because the architecture is composed of two separate networks; one is for classification and the other is for segmentation.
In the proposed network, object labels associated with an input image are identified by classification network while figure-ground segmentation of each identified label is subsequently obtained by segmentation network.
Additionally, there are bridging layers, which deliver class-specific information from classification to segmentation network and enable segmentation network to focus on the single label identified by classification network at a time.
%Decoupling classification and segmentation, training and testing of each network can be performed independently using image-level and pixel-wise annotations, respectively.

Training is performed on each network separately, where networks for classification and segmentation are trained with image-level and pixel-wise annotations, respectively; training does not require iterative procedure, and algorithm is easy to reproduce.
More importantly, decoupling classification and segmentation reduces search space for segmentation significantly, which makes it feasible to train the segmentation network with a handful number of segmentation annotations.
Inference in our network is also simple and does not involve any post-processing.
Extensive experiments show that our network substantially outperforms existing semi-supervised techniques based on DNNs even with much smaller segmentation annotations, {\it e.g.,} 5 or 10 strong annotations per class.

The rest of the paper is organized as follows. 
We briefly review related work and introduce overall algorithm in Section~\ref{sec:relatedwork} and \ref{sec:overview}, respectively.
The detailed configuration of the proposed network is described in Section~\ref{sec:architecture}, and  training algorithm is presented in Section~\ref{sec:training}. 
Section~\ref{sec:experiments} presents experimental results on a challenging benchmark dataset.

%% file: relatedwork.tex
% !TEX root = nips2015_transfer.tex

\section{Related Work}
\label{sec:relatedwork}
\ifdefined\paratitle{\color{blue}
[Supervised DNNs for semantic segmenation]
\\}\fi
Recent breakthrough in semantic segmentation are mainly driven by supervised approaches relying on Convolutional Neural Network (CNN)~\cite{Fcn,Deeplabcrf,Hypercolumns,Sds,Zoomout}.
Based on CNNs developed for image classification, they train networks to assign semantic labels to local regions within images such as pixels~\cite{Fcn,Deeplabcrf,Hypercolumns} or superpixels~\cite{Sds,Zoomout}.
Notably, Long \textit{et al.}~\cite{Fcn} propose an end-to-end system for semantic segmentation by transforming a standard CNN for classification into a fully convolutional network.
Later approaches improve segmentation accuracy through post-processing based on fully-connected CRF~\cite{Deeplabcrf,Crfrnn}.
Another branch of semantic segmentation is to learn a multi-layer deconvolution network, which also provides a complete end-to-end pipeline~\cite{deconvnet}. 
%However, training these networks requires images with pixel-wise ground-truths, and overall performance is heavily depended on size and quality of training data.
However, training these networks requires a large number of segmentation ground-truths, but the collection of such dataset is a difficult task due to excessive annotation efforts.

\iffalse
Based on the network proposed in \cite{Fcn}, Chen \textit{et al.} add the post-processing based on fully-connected CRF~\cite{Deeplabcrf} to obtain finer segmentation.
%On the other hand, Noh \textit{et al.} put deconvolution network on top of standard classification network and achieve 
On the other hand, Noh \textit{et al.} improve the segmentation accuracy by putting deconvolution network on top of standard classification network and performing instance-wise segmentation.
However, training these networks require notorious annotation efforts, since overall performance is heavily depended on size and quality of pixel-wise annotations.
\fi

\ifdefined\paratitle{\color{blue}
[Weakly-supervised DNNs for semantic segmenation]
\\}\fi
To mitigate heavy requirement of training data, weakly-supervised learning approaches start to draw attention recently.
In weakly-supervised setting, the models for semantic segmentation have been trained with only image-level labels~\cite{Wsl,Wssl,Fcmil} or bounding box class labels~\cite{Boxsup}. 
Given weakly annotated training images, they infer latent segmentation masks based on Multiple Instance Learning (MIL)~\cite{Wsl, Fcmil} or Expectation-Maximization (EM)~\cite{Wssl} framework based on the CNNs for supervised semantic segmentation. 
However, performance of weakly supervised learning approaches except \cite{Boxsup} is substantially lower than supervised methods, mainly because there is no direct supervision for segmentation during training.
Note that \cite{Boxsup} requires bounding box annotations as weak supervision, which are already pretty strong and significantly more expensive to acquire than image-level labels.
%However, recent approaches on weakly-supervised semantic segmentation tend be inaccurate compared to fully supervised approaches, mainly because there is no direct supervision for segmentation.

\ifdefined\paratitle{\color{blue}
[Semi-supervised DNNs for semantic segmenation]
\\}\fi
%Semi-supervised learning approaches, which assume that small number of segmentation masks are available in addition to large number of weak annotations, can be an alternative to resolve issues in supervised and weakly supervised approaches.
%we assume that small number of segmentation masks are available in addition to large number of weak annotations
Semi-supervised learning is an alternative to bridge the gap between fully- and weakly-supervised learning approaches.
In the standard semi-supervised learning framework, given only a small number of training images with strong annotations, one needs to infer the full segmentation labels for the rest of the data.
However, it is not plausible to learn a huge number of parameters in deep networks reliably in this scenario.
Instead, \cite{Wssl, Boxsup} train the models based on heterogeneous annotations---a large number of weak annotations as well as a small number strong annotations.
This approach is motivated from the facts that the weak annotations, {\it i.e.,} object labels per bounding box or image, is much more easily accessible than the strong ones and that the availability of the weak annotations is useful to learn a deep network by mining additional training examples with full segmentation masks.
Based on supervised CNN architectures, they iteratively infer and refine pixel-wise segmentation labels of weakly annotated images with guidance of strongly annotated images, where image-level labels~\cite{Wssl} and bounding box annotations~\cite{Boxsup} are employed as weak annotations. %region proposals~\cite{Boxsup}. 
They claim that exploiting few strong annotations substantially improves the accuracy of semantic segmentation while it reduces annotations efforts for supervision significantly.
However, they rely on iterative training procedures, which are often ad-hoc and heuristic and increase complexity to reproduce results in general.
Also, these approaches still need a fairly large number of strong annotations to achieve reliable performance.
%{\color{red}However, underlying model for most semi-supervised learning approaches rely on supervised CNNs, [I don't understand this sentence.]}

%% file: overview.tex
% !TEX root = nips2015_transfer.tex
\section{Algorithm Overview}
\label{sec:overview}
\begin{figure}[t!]
\centering
\includegraphics[width=1\linewidth] {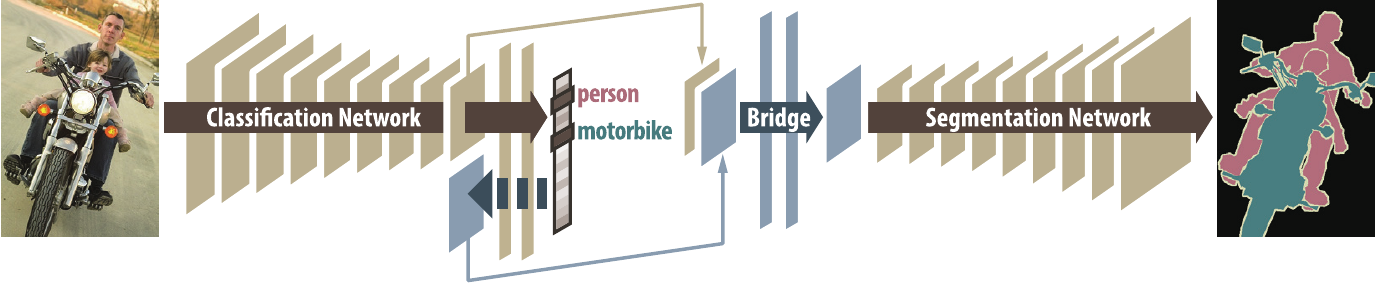}
\caption{The architecture of the proposed network. While classification and segmentation networks are decoupled, bridging layers deliver critical information from classification network to segmentation network.}
\label{fig:overall_network}
\end{figure}

\ifdefined\paratitle{\color{blue}[How we perform semantic segmentation using the proposed network]\\}\fi
%[What is the benefit of separating classification and segmentation network.]
Figure~\ref{fig:overall_network} presents the overall architecture of the proposed network.
Our network is composed of three parts: classification network, segmentation network and bridging layers connecting the two networks.
In this model, semantic segmentation is performed by separate but successive operations of classification and segmentation.
%given an input image, classification network identifies labels associated with image, and segmentation network subsequently produces figure-ground segmentation of each identified label. 
Given an input image, classification network identifies labels associated with the image, and segmentation network produces pixel-wise figure-ground segmentation corresponding to each identified label. 
%To design the segmentation network task-independent and transferrable between different domains, we add bridging layers between the two networks to deliver 	domain-specific information from classification to segmentation network. 
This formulation may suffer from loose connection between classification and segmentation, but we figure out this challenge by adding bridging layers between the two networks and delivering class-specific information from classification network to segmentation network. 
Then, it is possible to optimize the two networks using separate objective functions while the two decoupled tasks collaborate effectively to accomplish the final goal.
%Due to the connection using bridging layers, classification and segmentation networks can have separate objective functions and it enables separate optimization of the two networks.
%Due to the connection by bridging layers, objective of segmentation network becomes binary segmentation of the given class, and it enables independent optimization of the two networks.
%Note that bridging layers are key to decouple classification and segmentation network;
%it enables the two network to have separate objective functions---multi-class classification and binary segmentation---while achieving objective of standard semantic segmentation by combinations of the two networks.

%During training, classification and segmentation network is trained independently using image-level and pixel-wise annotations, respectively. 
\ifdefined\paratitle{\color{blue}[How to train the network]\\}\fi
%During training, we first train the classification network using image-level annotations. 
Training our network is very straightforward.
We assume that a large number of image-level annotations are available while there are only a few training images with segmentation annotations.
Given these heterogeneous and unbalanced training data, we first learn the classification network using rich image-level annotations. 
Then, with the classification network fixed, we jointly optimize the bridging layers and the segmentation network using a small number of training examples with strong annotations. 
There are only a small number of strongly annotated training data, but we alleviate this challenging situation by generating many artificial training examples through data augmentation.

\iffalse
Note that bridging layers play an important role in decoupling classification and segmentation network.
It delivers information for specific class from classification to segmentation network, and limit the objective of segmentation network as figure-ground segmentation of the given class.
This decoupling enables separate and independent training of classification and segmentation networks, which is advantageous to exploit heterogeneous annotations and reduce possible search space for segmentation.
\fi

\ifdefined\paratitle{\color{blue}[Contribution]\\}\fi
%This training strategy has the following advantages:
The contributions and characteristics of the proposed algorithm are summarized below:
\begin{itemize}[leftmargin=0.75cm]
\item We propose a novel DNN architecture for semi-supervised semantic segmentation using heterogeneous annotations. 
The new architecture decouples classification and segmentation tasks, which enables us to employ pre-trained models for classification network and train only segmentation network and bridging layers using a few strongly annotated data.
%\item We propose a novel DNN architecture for semi-supervised semantic segmentation using heterogeneous annotations. 
%The new architecture decouples classification and segmentation networks while bridging layers are employed to connect two networks effectively.
%
\item The bridging layers construct class-specific activation maps, which are delivered from classification network to segmentation network.
These maps provide strong priors for segmentation, and reduce search space dramatically for training and inference.
%\item Our algorithm reduces search space for inference and improves scalability in terms of the number of classes;
%we estimate binary figure-ground segmentation maps only for the relevant classes identified by classification network with guidance of class-specific saliency maps. 
%
\item Overall training procedure is very simple since two networks are to be trained separately.
Our algorithm does not infer segmentation labels of weakly annotated images through iterative heuristics\footnote{Due to this property, our framework is different from standard semi-supervised learning but close to few-shot learning with heterogeneous annotations. Nonetheless, we refer to it as {\em semi-supervised} learning in this paper since we have a fraction of strongly annotated data in our training dataset but complete annotations of weak labels. Note that our level of supervision is similar to the semi-supervised learning case in \cite{Wssl}.}, which are common in semi-supervised learning techniques~\cite{Wssl,Boxsup}.
\end{itemize} 
The proposed algorithm provides a concept to make up for the lack of strongly annotated training data using a large number of weakly annotated data.
This concept is interesting because the assumption about the availability of training data is desirable for real situations. 
We estimate figure-ground segmentation maps only for the relevant classes identified by classification network, which improves scalability of algorithm in terms of the number of classes. 
%It is possible to learn a new network for semantic segmentation given any classification network as long as a few training images with strong annotations per class are available.
Finally, our algorithm outperforms the comparable semi-supervised learning method with substantial margins in various settings.

%Details of each component of the network and procedures for applying the network are described in following subsections. 

\iffalse
\ifdefined\paratitle{\color{blue}[How to train the network in source and target domain]\\}\fi
During pre-training, we learn the entire network using any available source of pixel-wise annotations in fully-supervised manner. 
To transfer the network to the new domain, we fine-tune the network for classification and segmentation using images with class labels and pixel-wise annotations, respectively.
Since general knowledges for segmentation is transferred through fine-tuning of segmentation network, we show that entire network for semantic segmentation can be adapted to the new domain with surprisingly small number of pixel-wise annotations.
\fi

%% file: architecture_clssegbrg.tex
% !TEX root = nips2015_transfer.tex
\section{Architecture}
%\section{Stacked Classification and Segmentation Network for Semantic Segmentation}
\label{sec:architecture}

%In this section, we describe the detailed configuration of the proposed network, including classification network, segmentation network and a bridging layer enforcing the two networks to produce coherent predictions.
This section describes the detailed configurations of the proposed network, including classification network, segmentation network and bridging layers between the two networks.

\subsection{Classification Network}
\label{sec:classification}
\ifdefined\paratitle{\color{blue}
[input, output and objective function of the classification network.]
\\}\fi
%The classification network takes a raw image $\x$ as an input, and outputs likelihood vector $P(\x)\in\mathrm{R}^L$ indicating the probability of an image containing a subset of $L$ categories.
The classification network takes an image $\x$ as its input, and outputs a normalized score vector $S(\x ; \theta_c)\in\mathrm{R}^L$ representing a set of relevance scores of the input $\x$ based on the trained classification model $\theta_c$ for predefined $L$ categories.
The objective of classification network is to minimize error between ground-truths and estimated class labels, and is formally written as
%This can be achieved by minimizing sigmoid cross entropy between ground truth class label and predicted likelihood, which can be formally written by
%
\begin{equation}
\min_{\theta_c} \sum_i e_c(\y_i, S(\x_i ; \theta_c)),
\label{eq:obj_cls}
\end{equation}
where $\y_i\in\{ 0, 1\}^L$ denotes the ground-truth label vector of the $i$-th example and $e_c(\y_i, S(\x_i ; \theta_c))$ is classification loss of $S(\x_i ; \theta_c)$ with respect to $\y_i$.

\ifdefined\paratitle{\color{blue}
[Architecture of classification network]
\\}\fi
We employ VGG 16-layer net~\cite{Vgg16} as the base architecture for our classification network.
It consists of 13 convolutional layers, followed by rectification and optional pooling layers, and 3 fully connected layers for domain-specific projection. 
Sigmoid cross-entropy loss function is employed in Eq.~\eqref{eq:obj_cls}, which is a typical choice in  multi-class classification tasks.
%We employ the sigmoid cross entropy~\cite{entropy} as a loss function which is a typically choice in many multi-class classification task.
%We refer the readers to \cite{Vgg16} for more details of the network configuration.

Given output scores $S(\x_i ; \theta_c)$, our classification network identifies a set of labels $\mathcal{L}_i$ associated with input image $\x_i$. 
The region in $\x_i$ corresponding to each label $l \in \mathcal{L}_i$ is predicted by the segmentation network discussed next.
%Then, for each label $l \in \mathcal{L}_i$, corresponding region in $\x_i$ is predicted by the segmentation network, which will be discussed next.
%whose inputs are $7 \times 7$ class-specific activation map $\g^l$ obtained from bridging layers as well as $\mathsf{pool5}$ outputs from the classification network in the same size.
%In the following section, we first describe how segmentation network generates figure-ground segmentation of the class $l$ given $\g^l$, while details of constructing $\g^l$ in bridging layers are described in Section~\ref{sec:bridginglayers}.

\iffalse
Given output scores $P(\x_i)$, the classification network identifies a set of labels $\mathbf{l}^*$ associated with an input image. 
Then for each label $l\in \mathbf{l}^*$, the algorithm predicts corresponding regions in $\x_i$ by forward-propagating activations through segmentation network.
To this end, we employ \textit{bridging layers} to generate class-specific activation $\g^l$ from outputs of classification network, which will be served as inputs to segmentation network. 
In the following section, we first describe how segmentation network generates figure-ground segmentation of class $l$ given $\g^l$, while details of constructing $\g^l$ in bridging layers are described in Section~\ref{sec:bridginglayers}.
\fi

\subsection{Segmentation Network}
\label{sec:segmentation}
\ifdefined\paratitle{\color{blue}
[input, output and objective function of the segmentation network.]
\\}\fi
The segmentation network takes a class-specific activation map $\g_i^l$ of input image $\x_i$, which is obtained from bridging layers, and produces a two-channel class-specific segmentation map $M(\g_i^l ; \theta_s)$ after applying softmax function, where $\theta_s$ is the model parameter of segmentation network.
%The segmentation network takes a $7 \times 7$ class-specific activation map $\g_i^l$ of input image $\x_i$, which is obtained from bridging layers, and produces a two-channel segmentation map (after applying softmax function) $M(\g_i^l ; \theta_s)$ corresponding to class $l$, where $\theta_s$ is the model parameter of segmentation network.
Note that $M(\g_i^l ; \theta_s)$ has foreground and background channels, which are denoted by $M_f(\g_i^l ; \theta_s)$ and $M_b(\g_i^l ; \theta_s)$, respectively.
The segmentation task is formulated as per-pixel regression to ground-truth segmentation, which minimizes
\begin{equation}
\min_{\theta_s} \sum_i e_s(\z^l_i, M(\g_i^l ; \theta_s)),
\label{eq:obj_seg}
\end{equation}
where $\z^l_i$ denotes the binary ground-truth segmentation mask for category $l$ of the $i$-th image $\x_i$ and $e_s(\z_i, M(\g_i^l ; \theta_s))$ is the segmentation loss of $M_f(\g_i^l ; \theta_s)$---or equivalently the segmentation loss of $M_b(\g_i^l ; \theta_s)$---with respect to $\z^l_i$. 

\ifdefined\paratitle{\color{blue}
[Architecture of segmentation network]
\\}\fi
The recently proposed deconvolution network~\cite{deconvnet} is adopted for our segmentation network. 
Given an input activation map $\g^l_i$ corresponding to input image $\x_i$, the segmentation network generates a segmentation mask in the same size to $\x_i$ by multiple series of operations of unpooling, deconvolution and rectification. 
Unpooling is implemented by importing the switch variable from every pooling layer in the classification network, and the number of deconvolutional and unpooling layers are identical to the number of convolutional and pooling layers in the classification network.
We employ the softmax loss function to measure per-pixel loss in Eq.~\eqref{eq:obj_seg}.

\ifdefined\paratitle{\color{blue}
[What is the advantage of estimating binary segmentation]
\\}\fi
\iffalse
Note that our segmentation network predicts only \textit{binary} segmentation of the given class $l$, while the network proposed in \cite{deconvnet} predicts segmentations on all $L$ predefined classes. 
By decoupling classification from segmentation and limiting the objective of segmentation network as binary classification, it reduces number of parameters and search space for segmentation network significantly. 
This property is especially advantageous in a few-shot learning scenario, since only handful number of pixel-wise annotations (5 to 10 annotations per class) are available during training.
In following section, we describe how we encode class-specific information on the input to segmentation network, $\g^l$, by bridging layers.
\fi
%Note that our segmentation network predicts binary segmentation mask of the \textit{given} class $l$, while the network proposed in \cite{deconvnet} predicts segmentation masks of all  $L$ predefined classes. 
Note that the objective function in Eq.~\eqref{eq:obj_seg} corresponds to pixel-wise \textit{binary} classification; it infers whether each pixel belongs to the given class $l$ or not.
This is the major difference from the existing networks for semantic segmentation including \cite{deconvnet}, which aim to classify each pixel to one of the $L$ predefined classes.
By decoupling classification from segmentation and posing the objective of segmentation network as binary classification, our algorithm reduces the number of parameters in the segmentation network significantly. 
Specifically, this is because we identify the relevant labels using classification network and perform binary segmentation for each of the labels, where the number of output channels in segmentation network is set to two---for foreground and background---regardless of the total number of candidate classes.
This property is especially advantageous in our challenging scenario, where only a few pixel-wise annotations (typically 5 to 10 annotations per class) are available for training segmentation network.
%Training the deep network becomes much easier since solution space decreases significantly and half of the entire network is trained using rich image-level class annotations.
%This property is especially advantageous in our scenario, where only a few pixel-wise annotations (typically 5 to 10 annotations per class) are available while there are a sufficient number of training examples with image-level label annotations during training.

%To detach classification from segmentation, the inputs to segmentation network should encode information specific to class $l$. %as well as information for shape estimation.
%In the following section, we describe how such $\g^l$ is constructed by bridging layers.

\subsection{Bridging Layers}
\label{sec:bridginglayers}

\iffalse
As explained above, bridging layers are key to deliver information from classification to segmentation network.
It takes inputs from intermediate layers in classification network, and constructs class-specific activation map $\g^l$ for each identified label $l \in \mathbf{l}^*$ as an input to segmentation network. 
For accurate and class-specific segmentation, the class-specific activation map $\g^l$ should encode both spatial and class-specific information. 
\fi

To enable the segmentation network described in Section~\ref{sec:segmentation} to produce the segmentation mask of a specific class, the input to the segmentation network should involve class-specific information as well as spatial information required for shape generation.
To this end, we have additional layers underneath segmentation network, which is referred to as bridging layers, to construct the class-specific activation map $\g^l_i$ for each identified label $l \in \mathcal{L}_i$.

\ifdefined\paratitle{\color{blue}
[Encoding spatial information]
\\}\fi
%To impose spatial configuration of objects on $\g^l$, we adopt outputs from the last pooling layer in the classification network. 
To encode spatial configuration of objects presented in image, we exploit outputs from an intermediate layer in the classification network. 
We take the outputs from the last pooling layer ($\mathsf{pool5}$) since the activation patterns of convolution and pooling layers often preserve spatial information effectively while the activations in the higher layers tend to capture more abstract and global information.
We denote the activation map of $\mathsf{pool5}$ layer by $\f_{\text{spat}}$ afterwards.
%Although the feature is suitable to generate overall segmentation shape, however, it does not encode class-specific information.
%Regions from all know categories have high activation values in $\f_{\text{spat}}$, while there is no distinction about which activations corresponds to which categories.

\ifdefined\paratitle{\color{blue}
[Encoding class-specific information]
\\}\fi
%Although activations in $\f_{\text{spat}}$ preserve useful information for shape generation, it contains activations on all classes in $\mathbf{l}^*$ and cannot provide class-specific information.
%Although activations in $\f_{\text{spat}}$ preserve useful information for shape generation, it contains activations on all classes in $\mathbf{l}^*$ and we need additional information to identify relevance of activations in $\f_{\text{spat}}$ to specific class.
Although activations in $\f_{\text{spat}}$ maintain useful information for shape generation, they contain mixed information of all relevant labels in $\x_i$ and we should identify class-specific activations in $\f_{\text{spat}}$ additionally.
%To encode information on $\f_{\text{spat}}$, we need to identify relevance of activations in $\f_{\text{spat}}$ to specific class.
%Therefore, we employ class-specific saliency map proposed in \cite{saliency} as an additional input to bridging layer to encode class-specific information. 
%Therefore, we identify relevance of activations in $\f_{\text{spat}}$ to specific class using class-specific saliency map \cite{saliency}. 
For the purpose, we compute class-specific saliency maps using the back-propagation technique proposed in \cite{saliency}. 
Let $\f^{(i)}$ be the output of the $i$-th layer ($i = 1, \dots, M$) in the classification network. %, where $M$ denotes total number of layers in the classification network. 
The relevance of activations in $\f^{(k)}$ with respect to a specific class $l$ is computed by chain rule of partial derivative, which is similar to error back-propagation in optimization, as
\begin{equation}
\f^l_{\text{cls}} = \frac{\partial S_l}{\partial \f^{(k)}} = \frac{\partial \f^{(M)}}{\partial \f^{(M-1)}}\frac{\partial \f^{(M-1)}}{\partial \f^{(M-2)}} \cdots \frac{\partial \f^{(k+1)}}{\partial \f^{(k)}},
\label{eq:backpropagation}
\end{equation}
where $\f^l_{\text{cls}}$ denotes class-specific saliency map and $S_l$ is the classification score of class $l$. 
Intuitively, Eq.~\eqref{eq:backpropagation} means that the values in $\f^l_{\text{cls}}$ depend on how much the activations in $\f^{(k)}$ are relevant to class $l$; this is measured by computing the partial derivative of class score $S_l$ with respect to the activations in $\f^{(k)}$.
%{\color{red}Intuitively, Eq.~\eqref{eq:backpropagation} means that, as activations in $\f^{(k)}$ are more relevant to class $l$, they affect changes in class score $S_l$ more, which would result in high values in $\f^l_{\text{cls}}$.}
We back-propagate the class-specific information until $\mathsf{pool5}$ layer.
%which is the $k$-th layer in the classification network.
%We compute class-specific saliency map $\f^l_{\text{cls}}$ until the last pooling layer to discover importance of activations w.r.t. class $l$ in $\f_{\text{spat}}$.

\ifdefined\paratitle{\color{blue}
[Combining spatial and class-specific information to obtain inputs to segmentation network]
\\}\fi
%Given $\f_{\text{spat}}$ and $\f^l_{\text{cls}}$ obtained above, the class-specific activation $\g^l$ is computed by combining both activation maps.
The class-specific activation map $\g^l_i$ is obtained by combining both $\f_{\text{spat}}$ and $\f^l_{\text{cls}}$.
%Activations from the last pooling layer encode spatial location and patterns of all classes in $\mathbf{l}^*$, and the class-specific saliency map reveals importance of each activations in $\f^(k)$ for each class $l \in \mathbf{l}^*$. 
%To make $\g^l$ encode both information, we need to combine both information.
We first concatenate $\f_{\text{spat}}$ and $\f^l_{\text{cls}}$ in their channel direction, and forward-propagate it through the fully-connected bridging layers, which discover the optimal combination of $\f_{\text{spat}}$ and $\f^l_{\text{cls}}$ using the trained weights. 
%By training the network to find optimal $\g^l$, the combination maximizing the segmentation accuracy is discovered by weights in fully-connected layers.
%Note that the network discovers the optimal combination of $\f_{\text{spat}}$ and $\f^l_{\text{cls}}$ using the trained weights in the bridging layers. 
The resultant class-specific activation map $\g^l_i$ that contains both spatial and class-specific information is given to segmentation network to produce a class-specific segmentation map. 
Note that the changes in $\g^l_i$ depend only on $\f^l_{\text{cls}}$ since $\f_{\text{spat}}$ is fixed for all classes in an input image.

\ifdefined\paratitle {\color{blue}
[Effect of class-specific activation map $\g^l$ for better generalization of segmentation network]\\
}\fi
\begin{figure}[t!]
\centering
\includegraphics[width=1\linewidth] {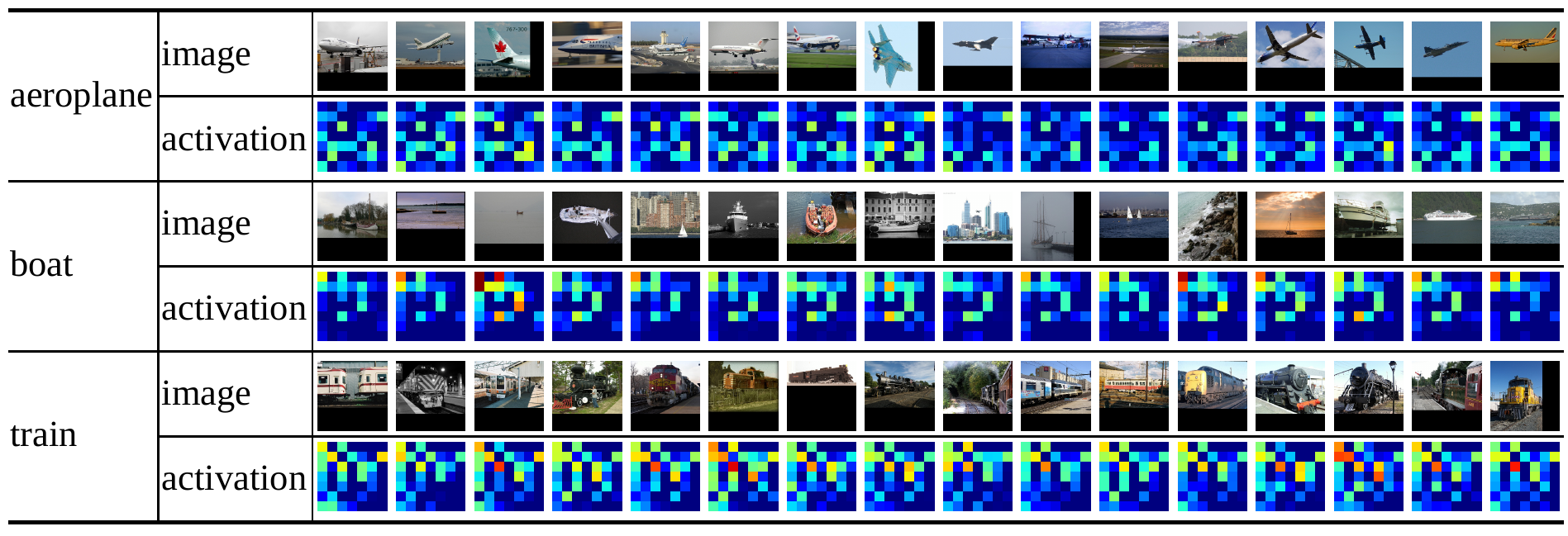}
%\vspace{-0.7cm}
\caption{Examples of class-specific activation maps (output of bridging layers). Despite significant variations in input images, the class-specific activation maps share similar properties, which suggests that the search space for segmentation may not be as huge as our prejudice.}
\label{fig:clssaliency_vis}
\end{figure}
Figure~\ref{fig:clssaliency_vis} visualizes the examples of class-specific activation maps $\g^l_i$ obtained from several validation images.
The activations from the images in the same class share similar patterns despite substantial appearance variations, which shows that the outputs of bridging layers capture class-specific information effectively; this property makes it possible to obtain figure-ground segmentation maps for individual relevant classes in segmentation network.
More importantly, it reduces the variations of input distributions for segmentation network, which allows to achieve good generalization performance in segmentation even with a small number of training examples.

%It is because $\g^l$ is constructed using outputs from intermediate layers in the classification network; information for each input example is abstracted in higher layers effectively.
%This is an important property since $\g^l$ is served as an input to segmentation network.
%By reducing variations of input activations, we can achieve excellent generalization performance even with small number of training examples for segmentation network.

%Since we compute $\g^l$ based on outputs from intermediate layers in the classification network, information of 
%Since we train the segmentation network with only small number of training examples, there is possible overfitting problem.
  
\begin{figure}[t!] \small
%\centering
%\includegraphics[width=1\linewidth] {../figures/seg_outputs.pdf}
\includegraphics[width=1\linewidth] {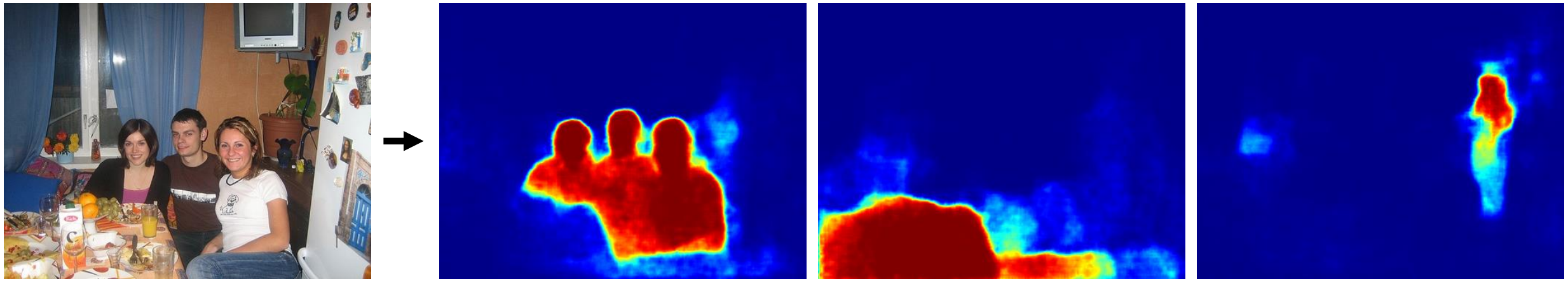}\\
\hspace{-0.05cm}$\mathcal{L}^*_i=\{\text{person}, \text{table}, \text{plant}\}$\hspace{1.4cm}
$M_f(\g_i^{\text{person}})$\hspace{1.8cm}
$M_f(\g_i^{\text{table}})$\hspace{1.9cm}
$M_f(\g_i^{\text{plant}})$
%\vspace{-0.7cm}
\caption{Input image (left) and its segmentation maps (right) of individual classes.}
\label{fig:seg_results}
\end{figure}

\ifdefined\paratitle {\color{blue}
[Applying class-specific activations to class-specific segmentation]\\
}\fi
For inference, we compute a class-specific activation map $\g^l_i$ for each identified label $l \in \mathcal{L}_i$ and obtain class-specific segmentation maps $\{ M(\g_i^l ; \theta_s) \}_{\forall l\in\mathcal{L}_i}$. 
In addition, we obtain $M(\g_i^* ; \theta_s)$, where $\g_i^*$  is the activation map from the bridging layers for all identified labels.
%Then, the final label estimation is given by identifying the label with the maximum score for each pixel out of all the class-specific foreground segmentation maps and the overall background segmentation map.
The final label estimation is given by identifying the label with the maximum score in each pixel out of $\{ M_f(\g_i^l ; \theta_s) \}_{\forall l\in\mathcal{L}_i}$ and $ M_b(\g_i^* ; \theta_s)$.
%
\iffalse
\begin{equation}
\max \left[ \left( \max_l M_f(\g_i^l ; \theta_s) \right), M_b(\g_i^\text{all} ; \theta_s) \right].
\end{equation}
\fi
%
Figure~\ref{fig:seg_results} illustrates the output segmentation map of each $\g^l_i$ for $\x_i$, where each map identifies high response area given $\g^l_i$ successfully.
%Note that outputs of the segmentation network changes depending on which class is used to compute $\g^l$, while all parameters in the network remain same.
%

%% file: trainAndInference.tex
% !TEX root = nips2015_transfer.tex

\section{Training}
\label{sec:training}

%We discuss training procedure of the proposed deep network and data augmentation technique to learn a reliable model in this section.

%\paragraph{Training classification and segmentation networks}
%In our semi-supervised learning scenario, we assume that we have combined datasets with weak and strong annotations.
In our semi-supervised learning scenario, we have mixed training examples with weak and strong annotations.
Let $\mathcal{W}=\{1,...,N_w\}$ and $\mathcal{S}=\{1,...,N_s \}$ denote the index sets of images with image-level and pixel-wise class labels, respectively, where $N_w \gg N_s$. 
We first train the classification network using the images in $\mathcal{W}$ by optimizing the loss function in Eq.~\eqref{eq:obj_cls}.
Then, fixing the weights in the classification network, we jointly train the bridging layers and the segmentation network using images in $\mathcal{S}$ by optimizing Eq.~\eqref{eq:obj_seg}.
For training segmentation network, we need to obtain class-specific activation map $\g^l_i$ from bridging layers using ground-truth class labels associated with $\x_i$, $i \in \mathcal{S}$.
Note that we can reduce complexity in training by optimizing the two networks separately.

%\paragraph{Data augmentation with segmentation labels} 
%Compared to images in set $A$, number of training examples for segmentation is too small.
Although the proposed algorithm has several advantages in training segmentation network with few training images, it would still be better to have more training examples with strong annotations.
%the number of training examples with strong annotations is still important to achieve good performance.
Hence, we propose an effective data augmentation strategy, \textit{combinatorial cropping}.
%we first construct all possible combinations of class labels associated with each training image. 
Let $\mathcal{L}^*_i$ denotes a set of ground-truth labels associated with image $\x_i, i \in \mathcal{S}$.
We enumerate all possible combinations of labels in $\mathsf{P}(\mathcal{L}^*_i)$, where $\mathsf{P}(\mathcal{L}^*_i)$ denotes the powerset of $\mathcal{L}^*_i$.
For each $\mathcal{P} \in \mathsf{P}(\mathcal{L}_i^*)$ except empty set ($\mathcal{P} \neq \emptyset$), we construct a binary ground-truth segmentation mask $\z_i^\mathcal{P}$ by setting the pixels corresponding to every label $l \in \mathcal{P}$ as foreground and the rests as background.
%Then we randomly crop multiple sub-images containing the foreground segmentations using region proposals~\cite{Edgebox}. 
%Since $\left| \Gamma_i \right| = 2^{\left|\mathbf{l^*_i}\right|} - 1$, the number of examples increases drastically and we can augment number of strong annotations effectively.
Then, we generate $N_p$ sub-images enclosing the foreground areas based on region proposal method~\cite{Edgebox} and random sampling.
Through this simple data augmentation technique, we have $N_t = N_s + N_p \cdot \left( \sum_{i\in \mathcal{S}} 2^{\left|\mathcal{L}^*_i\right|} - 1\right)$ training examples with strong annotations effectively, where $N_t \gg N_s$.

%For an image with $n$ associated labels, there are $\sum_{r=1}^n{{n}\choose{r}}=2^n-1$ combinations. 
%For each combination, we set regions in image corresponding to selected labels as foreground and the rests as background.
%For each combination, we construct ground-truth segmentation masks $\z_i^l$ by setting segmentations corresponding to class $l$ as foreground and the rests as background.
%where candidate windows for cropping are obtained by object proposal algorithms~\cite{edgebox}. 
%This procedure is effective to augment number of training examples, especially when only small number of images with segmentation annotations are available.

%% file: experiments.tex
% !TEX root = nips2015_transfer.tex
\section{Experiments}
\label{sec:experiments}

%This section describes implementation details and extensive evaluation results of the proposed algorithm.

%This section first describes our implementation details and experiment setup. 
%Then, we provide analysis and evaluation of the proposed network using standard benchmark dataset.

\subsection{Implementation Details}

\paragraph{Dataset}
We employ PASCAL VOC 2012 dataset~\cite{Pascalvoc} for training and testing of the proposed deep network.
The dataset with extended annotations from \cite{Hariharan}, which contains 12,031 images with pixel-wise class labels, is used in our experiment.
To simulate semi-supervised learning scenario, we divide the training images into two non-disjoint subsets---$\mathcal{W}$ with weak annotations only and $\mathcal{S}$ with strong annotations as well.
There are 10,582 images with image-level class labels, which are used to train our classification network.
%The subset with weak annotations contains 10,581 training images using only per-image class labels as ground-truth annotations.
%Then we construct a set of strong annotations by randomly selecting a small subset of training images annotated with pixel-wise class labels.
%A set of strong annotations are constructed with much smaller subset of training images using pixel-wise annotations.
We also construct training datasets with strongly annotated images; the number of images with segmentation labels per class is controlled to evaluate the impact of supervision level.
In our experiment, three different cases---5, 10, or 25 training images with strong annotations per class---are tested to show the effectiveness of our semi-supervised framework.
%we construct various sets of strong annotations by increasing numbers of examples per class by 5, 10, and 25.
%images with weak annotations with 10.581 training images, while constructing images with strong annotations by controlling number of examples per class by 5, 10, 25. 
We evaluate the performance of the proposed algorithm on 1,449 validation images.

%\paragraph{Training Data Construction}
\paragraph{Data Augmentation}
We employ different strategies to augment training examples in the two datasets with weak and strong annotations.
For the images with weak annotations, simple data augmentation techniques such as random cropping and horizontal flipping are employed as suggested in \cite{Vgg16}.
%For the images with weak annotations, simple data augmentation techniques are adopted as in \cite{Vgg16}, where random $320\times320$ sub-images are sampled from $350\times350$ resized input image and flipped horizontally.
%a simple technique proposed in~\cite{Vgg16} is used; we first resize an input image to $250\times250$ square image, and randomly sample $224\times224$ sub-images with horizontal flipping.
We perform combinatorial cropping proposed in Section~\ref{sec:training} for the images with strong annotations, where EdgeBox~\cite{Edgebox} is adopted to generate region proposals and the  $N_p (=200)$ sub-images are generated for each label combination.

\paragraph{Optimization}
We implement the proposed network based on Caffe library~\cite{caffe}. 
The standard Stochastic Gradient Descent (SGD) with momentum is employed for optimization, where all parameters are identical to \cite{deconvnet}.
We initialize the weights of the classification network using VGG 16-layer net pre-trained on ILSVRC~\cite{imagenet} dataset.
When we train the deep network with full annotations, the network converges after approximately 5.5K and 17.5K SGD iterations with mini-batches of 64 examples in training classification and segmentation networks, respectively; training takes 3 days (0.5 day for classification network and 2.5 days for segmentation network) in a single Nvidia GTX Titan X GPU with 12G memory.
Note that training segmentation network is much faster in our semi-supervised setting while there is no change in training time of classification network.

\subsection{Results on PASCAL VOC Dataset}
\iffalse
Things to discuss:
0. experiment setting: our method--decoupledNet. we compared the performance of DeconvNet, which has similar architecture based on deconvolution network for semantic segmentation. Since this algorithm cannot exploit weak annotations, we train the method with only segmentation annotations.
1-1. our method performs well even with a small number of segmentation annotations. considering difference of annotation size, the performance gap between fully supervised and semi supervised approach is not significant.
1-2. our method substantially outperforms other semi-supervised approach. the performance of our method is higher even with much smaller number of supervision. it is probably because decoupling of classification from segmentation reduces search space for segmentation network and enables effective learning of segmentation network with handful number of supervised examples.
1-3. To assess effectiveness of exploiting weak annotations, we compared performance of our method by training both classification and segmentation network using examples with strong annotations (without weak annotations). The performance drops significantly, which shows the effectiveness of exploiting weak annotations.
1-4. 
1-5. qualitative results.
\fi

%%%%%%%%%%%%%%%%%%%%%%%% INSERT TABLE HERE!
\begin{table*}[!t] \scriptsize
\centering
\caption{Evaluation results on PASCAL VOC 2012 validation set. %(Asterisk (*) denotes the algorithms with additional post-processing.)
} \vspace{0.1cm}%\vspace{-0.2cm}
\begin{tabular}
{c|ccc|cc}
%{
%@{}C{2.7cm}@{}|@{}C{0.68cm}@{}C{0.66cm}@{}C{0.66cm}@{}C{0.66cm}@{}C{0.66cm}@{}
%}
%\hline
%&DecoupleNet&$\text{DecoupleNet}^{\text{str}}$&DeconvNet~\cite{deconvnet}&WSSL~\cite{Wssl}*&WSL~\cite{Wsl}*\\
\# of strongs&DecoupledNet&WSSL-Small\_FoV~\cite{Wssl}&WSSL-Large-FoV~\cite{Wssl}&\text{DecoupledNet}-{\text{Str}}&DeconvNet~\cite{deconvnet}\\
\hline
Full&67.5&63.9&\textbf{67.6}&67.5&67.1\\
\hline
25 ($\times$20 classes)&\textbf{62.1}&56.9&54.2&50.3&38.6\\
10 ($\times$20 classes)&\textbf{57.4}&47.6&38.9&41.7&21.5\\
~5~ ($\times$20 classes)&\textbf{53.1}&-&-&32.7&15.3\\
\hline
\end{tabular}
\label{tab:voc_result}
\end{table*}
%%%%%%%%%%%%%%%%%%%%%%%%%%%%%%%%%%%%%%%%%%

%%%%%%%%%%%%%%%%%%%%%%%% INSERT TABLE HERE!
\begin{table*}[!t] \scriptsize
%\begin{table*}[!t] \scriptsize
\centering
\caption{Evaluation results on PASCAL VOC 2012 test set.} \vspace{0.1cm}%\vspace{-0.2cm}
\begin{tabular}
{
@{}C{2.0cm}@{}|@{}C{0.53cm}@{}C{0.54cm}@{}C{0.53cm}@{}C{0.53cm}@{}C{0.53cm}@{}C{0.565cm}@{}C{0.53cm}@{}C{0.53cm}@{}C{0.53cm}@{}C{0.53cm}@{}C{0.53cm}@{}C{0.53cm}@{}C{0.53cm}@{}C{0.57cm}@{}C{0.53cm}@{}C{0.53cm}@{}C{0.53cm}@{}C{0.57cm}@{}C{0.53cm}@{}C{0.53cm}@{}C{0.52cm}@{}|@{}C{0.56cm}@{}
}
%\hline
Models&bkg&areo&bike&bird&boat&bottle&bus&car&cat&chair&cow&table&dog&horse&mbk&prsn&plnt&sheep&sofa&train&tv&mean\\
\hline
%Hypercolumn~\cite{Hypercolumns}&88.9&68.4&27.2&68.2&47.6&61.7&76.9&72.1&71.1&24.3&59.3&44.8&62.7&59.4&73.5&70.6&52.0&63.0&38.1&60.0&54.1&59.2\\
%MSRA-CFM~\cite{DAICVPR15}&87.7&75.7&26.7&69.5&48.8&65.6&81.0&69.2&73.3&30.0&68.7&51.5&69.1&68.1&71.7&67.5&50.4&66.5&44.4&58.9&53.5&61.8\\
%FCN8s~\cite{Fcn}&91.2&76.8&34.2&68.9&49.4&60.3&75.3&74.7&77.6&21.4&62.5&46.8&71.8&63.9&76.5&73.9&45.2&72.4&37.4&70.9&55.1&62.2\\
%TTI-Zoomout-16~\cite{Zoomout}&89.8&81.9&35.1&78.2&57.4&56.5&80.5&74.0&79.8&22.4&69.6&53.7&74.0&76.0&76.6&68.8&44.3&70.2&40.2&68.9&55.3&64.4\\
%DeepLab-CRF~\cite{Deeplabcrf}&92.1&78.4&33.1&78.2&55.6&65.3&81.3&75.5&78.6&25.3&69.2&52.7&75.2&69.0&79.1&77.6&54.7&78.3&45.1&73.3&56.2&66.4\\
%
%%%% DecoupledNet - full
%DecoupledNet-Full&\textbf{91.5} &\textbf{78.8} &39.9 &\textbf{78.1} &\textbf{53.8} &\textbf{68.3} &\textbf{83.2} &\textbf{78.2} &\textbf{80.6} &\textbf{25.8} &\textbf{62.6} &\textbf{55.5} &\textbf{75.1} &\textbf{77.2} &\textbf{77.1} &\textbf{76.0} &\textbf{47.8} &\textbf{74.1} &\textbf{47.5} &\textbf{66.4} &\textbf{60.4}&\textbf{66.6} \\
%\hline
DecoupledNet-Full&91.5 &78.8 &39.9 &78.1 &53.8 &68.3 &83.2 &78.2 &80.6 &25.8 &62.6 &55.5 &75.1 &77.2 &77.1 &76.0 &47.8 &74.1 &47.5 &66.4 &60.4&66.6 \\
\hline
%%%% DecoupledNet - 25
DecoupledNet-25&90.1 &75.8 &41.7 &70.4 &46.4 &66.2 &83.0 &69.9 &76.7 &23.1 &61.2 &43.3 &70.4 &75.7 &74.1 &65.7 &46.2 &73.8 &39.7 &61.9 &57.6&62.5 \\

%%%% DecoupledNet - 10
DecoupledNet-10&88.5 &73.8 &40.1 &68.1 &45.5 &59.5 &76.4 &62.7 &71.4 &17.7 &60.4 &39.9 &64.5 &73.0 &68.5 &56.0 &43.4 &70.8 &37.8 &60.3 &54.2&58.7 \\

%%%% DecoupledNet - 5
DecoupledNet-5&87.4 &70.4 &40.9 &60.4 &36.3 &61.2 &67.3 &67.7 &64.6 &12.8 &60.2 &26.4 &63.2 &69.6 &64.8 &53.1 &34.7 &65.3 &34.4 &57.0 &50.5&54.7 \\
%%%% DecoupledNet-Str-25

%%%% DecoupledNet-Str-10

%%%% DecoupledNet-Str-5

%* WSSL~\cite{Weaklyandsemi}&93.2&85.3&36.2&\bf{84.8}&61.2&67.5&84.7&81.4&81.0&\bf{30.8}&73.8&53.8&77.5&76.5&82.3&\bf{81.6}&56.3&78.9&52.3&76.6&63.3&70.4\\
%* BoxSup~\cite{Boxsup}&\bf{93.6}&86.4&35.5&79.7&\bf{65.2}&65.2&84.3&78.5&83.7&30.5&76.2&\bf{62.6}&\bf{79.3}&76.1&82.1&81.3&57.0&78.2&\bf{55.0}&72.5&\bf{68.1}&71.0\\
%
\hline
\end{tabular}
\label{tab:voc_result_test}
\end{table*}
%%%%%%%%%%%%%%%%%%%%%%%%%%%%%%%%%%%%%%%%%%

\ifdefined\paratitle{\color{blue}
[Brief descriptions about experiment settings, name of method, reference to table]
\\}\fi
Our algorithm denoted by DecoupledNet is compared with two variations in WSSL~\cite{Wssl}, which is another algorithm based on semi-supervised learning with heterogeneous annotations. 
We also test the performance of DecoupledNet-Str\footnote{This is identical to DecoupledNet except that its classification and segmentation networks are trained with the same images, where image-level weak annotations are generated from pixel-wise segmentation annotations.} and DeconvNet~\cite{deconvnet}, which only utilize examples with strong annotations, to analyze the benefit of image-level weak annotations.
%We also test the performance of DecoupledNet-Str\footnote{This is identical to DecoupledNet except that its classification and segmentation networks are trained with the same data, where examples with weak annotations are generated from strongly annotated ones.} and DeconvNet~\cite{deconvnet}, which only utilize examples with strong annotations, to analyze the impact of image-level weak annotations.
%share same architecture with only difference in training data.
%\footnote{DecoupledNet-Str is our method training both classification and segmentation networks with a small number of strong annotations.} 
All learned models in our experiment are based only on the training set (not including the validation set) in PASCAL VOC 2012 dataset.
All algorithms except WSSL~\cite{Wssl} report the results without CRF.
Segmentation accuracy is measured by Intersection over Union (IoU) between ground-truth and predicted segmentation, and the mean IoU over 20 semantic categories is employed for the final performance evaluation.

Table~\ref{tab:voc_result} summarizes quantitative results on PASCAL VOC 2012 validation set.
Given the same amount of supervision, DecoupledNet presents substantially better performance even without any post-processing than WSSL~\cite{Wssl}, which is a directly comparable method.
In particular, our algorithm has great advantage over WSSL when the number of strong annotations is extremely small.
We believe that this is because DecoupledNet reduces search space for segmentation effectively by employing the bridging layers and the deep network can be trained with a smaller number of images with strong annotations consequently.
Our results are even more meaningful since training procedure of DecoupledNet is very straightforward compared to WSSL and does not involve heuristic iterative procedures, which are common in semi-supervised learning methods.

When there are only a small number of strongly annotated training data, our algorithm obviously outperforms DecoupledNet-Str and DeconvNet~\cite{deconvnet} by exploiting the rich information of weakly annotated images.
It is interesting that DecoupledNet-Str is clearly better than DeconvNet, especially when the number of training examples is small.
%The benefit of using extra weak annotations is clearly observed from the performance gap between DeconvNet and DeconvNet-Str, where weak annotations improve the performance significantly especially with a small number of strong annotations.
For reference, the best accuracy of the algorithm based only on the examples with image-level labels is 42.0\%~\cite{Wsl}, which is much lower than our result with five strongly annotated images per class, even though \cite{Wsl} requires significant efforts for heuristic post-processing.
These results show that even little strong supervision can improve semantic segmentation performance dramatically.

Table~\ref{tab:voc_result_test} presents more comprehensive results of our algorithm in PASCAL VOC test set.
Our algorithm works well in general and approaches to the empirical upper-bound fast with a small number of strongly annotated images.
%while it still has potential for further improvement by including the validation set for training.
%
\ifdefined\paratitle{\color{blue}
[comparison to supervised method and benefit of using weak anotations]
\\}\fi
A drawback of our algorithm is that it does not achieve the state-of-the-art performance \cite{Deeplabcrf,Crfrnn,deconvnet} when the (almost\footnote{We did not include the validation set for training and have less training examples  than the competitors.}) full supervision is provided in PASCAL VOC dataset.
This is probably because our method optimizes classification and segmentation networks separately although joint optimization of two objectives is more desirable.
However, note that our strategy is more appropriate for semi-supervised learning scenario as shown in our experiment.

\iffalse
DecoupleNet and DecoupleNet-Str denote our method trained with full annotations (weak+strong) and only strong annotations, respectively.
DeconvNet~\cite{deconvnet} is supervised method having similar architecture with the proposed network based on deconvolution network.
WSSL~\cite{Wssl} is a semi-supervised approach most relevant to ours, which exploits both weakly and strong annotations for semantic segmentation. 
WSL~\cite{Wsl} is weakly-supervised semantic segmentation method based on DNN.
%To provide thorough analysis of the proposed algorithm, we compared our method with state-of-the-art DNN-based methods for fully-, semi- and weakly-supervised approaches denoted by DeconvNet~\cite{deconvnet}, WSSL~\cite{Wssl} and WSL~\cite{Wsl}, respectively.
To better understand the proposed algorithm, we compare it with approaches based on different levels of supervisions such as fully-supervised (DeconvNet~\cite{deconvnet}), semi-supervised (WSSL~\cite{Wssl}) and weakly-supervised  (WSL~\cite{Wsl}) approaches for semantic segmentation.
%DeconvNet~\cite{deconvnet}, WSSL~\cite{Wssl} and WSL~\cite{Wsl} is DNN-based method for supervised, semi-supervised, and weakly-supervised semantic segmentation, respectively.
\fi

\ifdefined\paratitle{\color{blue}
[effect of number of supervision to the performance]
\\}\fi
\iffalse
To assess effect of strong supervision, we conduct several experiments by controlling number of strong annotations per each category to 5, 10, and 25. 
%We first provides internal analysis of the proposed algorithm.
As the number of strong annotations increases, performance of all methods improves while the proposed DecoupleNet substantially outperforms all others.
our method performs well even with a small number of segmentation annotations. considering difference of annotation size, the performance gap between fully supervised and semi supervised approach is not significant.

The performance of supervised approach~\cite{deconvnet} degrades significantly in semi-supervised setting, since they can only exploit a small number of strong annotations while our method can benefit from a large number of weak annotations.
The benefit of using extra weak annotations is clearly observed from the performance gap between DeconvNet and DeconvNet-Str, where weak annotations improve the performance significantly especially with a small number of strong annotations.
\fi

\ifdefined\paratitle{\color{blue}
[comparison to semi-supervised method: WSSL]
\\}\fi
\iffalse
Compared to semi-supervised approach~\cite{Wssl}, our method performs substantially better given a same amount of strong supervision, and even with smaller examples. 
It is mainly because the performance of WSSL degrades significantly as number of strong annotations decreases, while relative performance drop of our method is not significant.
It is due to 
\fi

%our method substantially outperforms other semi-supervised approach. the performance of our method is higher even with much smaller number of supervision. it is probably because decoupling of classification from segmentation reduces search space for segmentation network and enables effective learning of segmentation network with handful number of supervised examples.

\ifdefined\paratitle{\color{blue}
[comparison to weakly-supervised method: WSL, with performance in test dataset]
\\}\fi

\begin{figure}[!t]
\centering
\vspace{-0.4cm}
\includegraphics[width=0.98\linewidth] {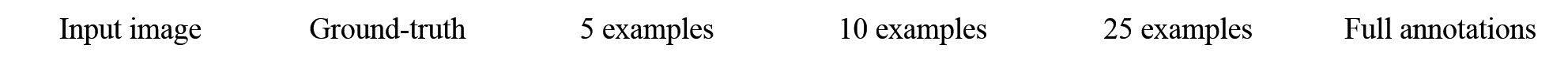}\\ \vspace{-0.13cm}
\begin{minipage}{1\textwidth}
\centering
\includegraphics[width=0.16\linewidth] {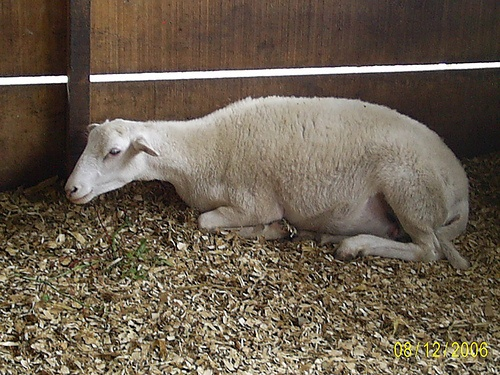}
\includegraphics[width=0.16\linewidth] {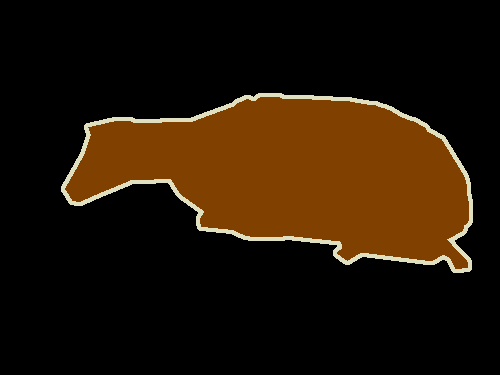}
\includegraphics[width=0.16\linewidth] {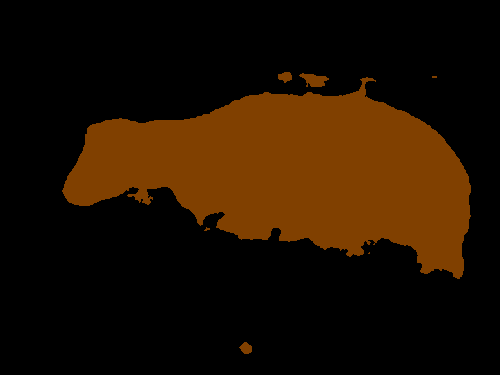}
\includegraphics[width=0.16\linewidth] {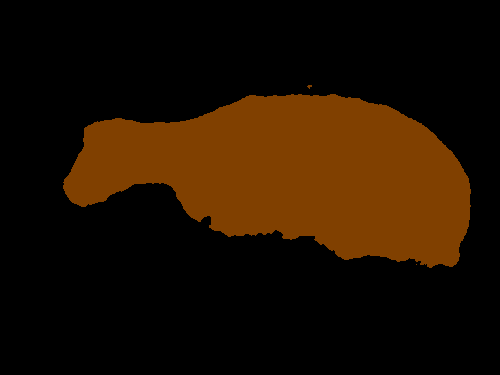}
\includegraphics[width=0.16\linewidth] {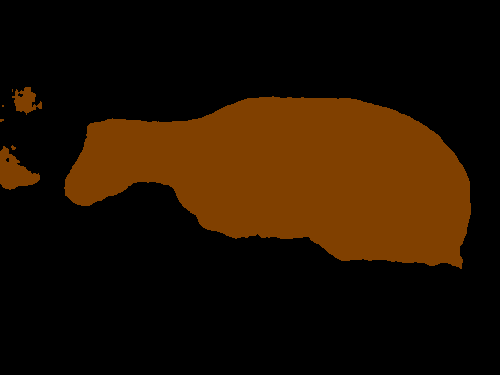}
\includegraphics[width=0.16\linewidth] {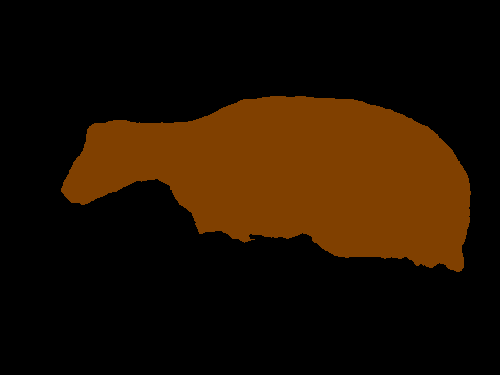}  \\ 
\includegraphics[width=0.16\linewidth] {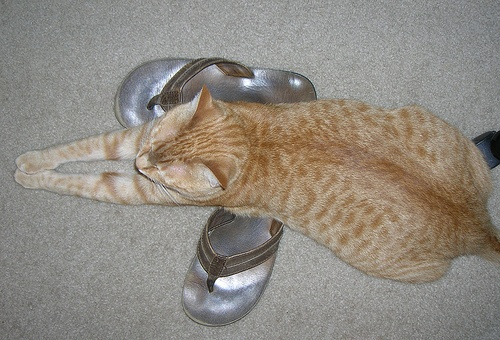}
\includegraphics[width=0.16\linewidth] {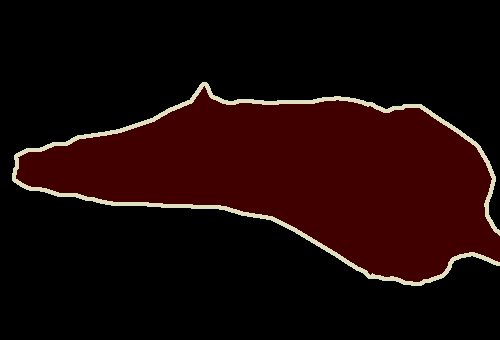}
\includegraphics[width=0.16\linewidth] {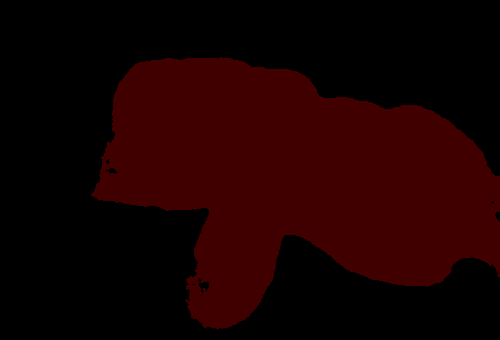}
\includegraphics[width=0.16\linewidth] {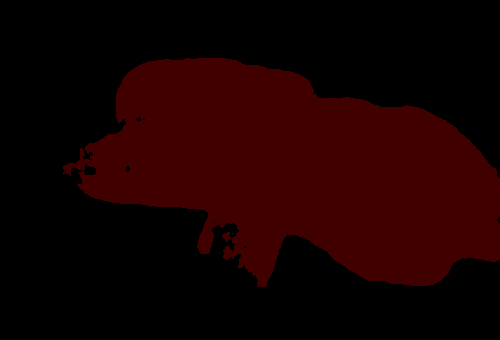}
\includegraphics[width=0.16\linewidth] {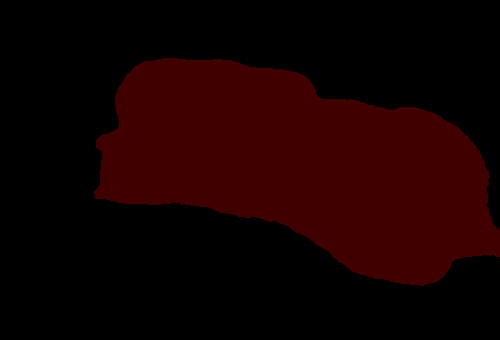}
\includegraphics[width=0.16\linewidth] {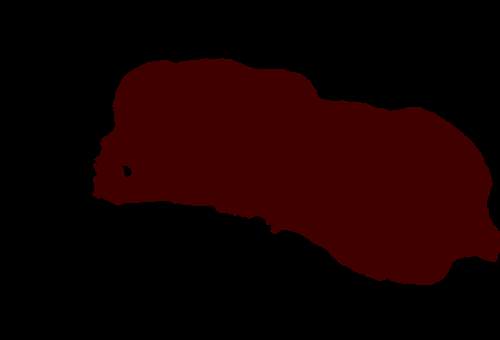}  \\ 
\includegraphics[width=0.16\linewidth] {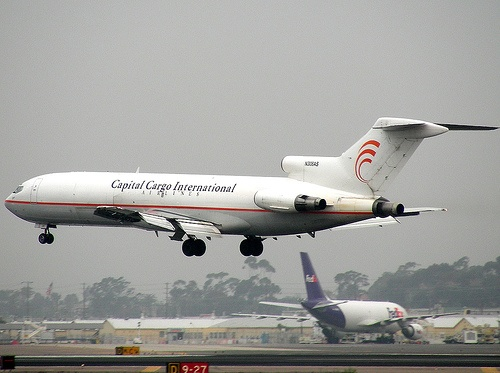}
\includegraphics[width=0.16\linewidth] {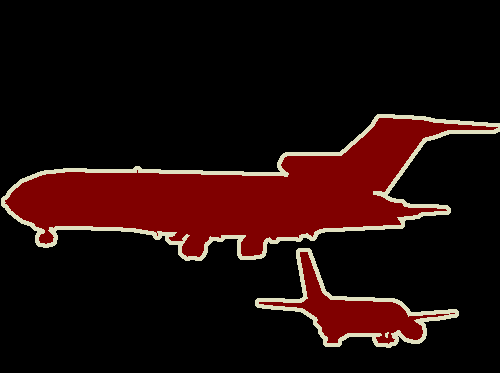}
\includegraphics[width=0.16\linewidth] {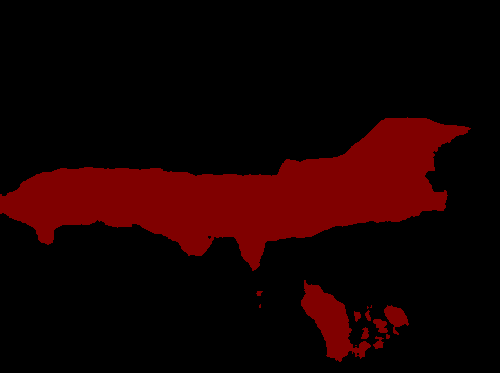}
\includegraphics[width=0.16\linewidth] {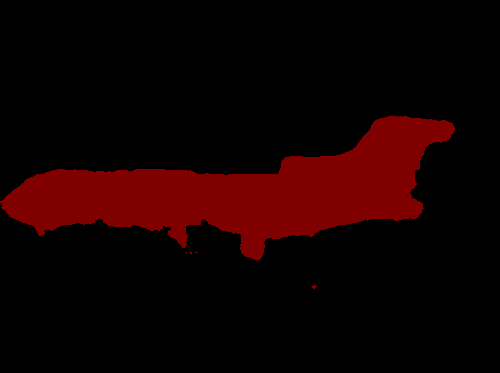}
\includegraphics[width=0.16\linewidth] {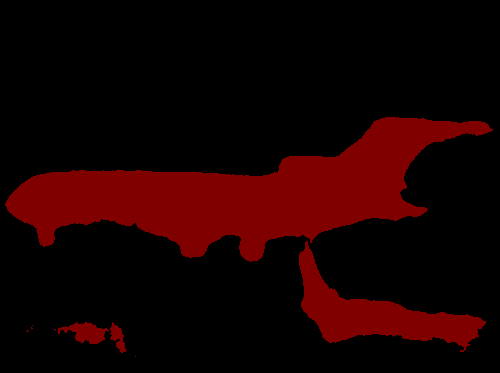}
\includegraphics[width=0.16\linewidth] {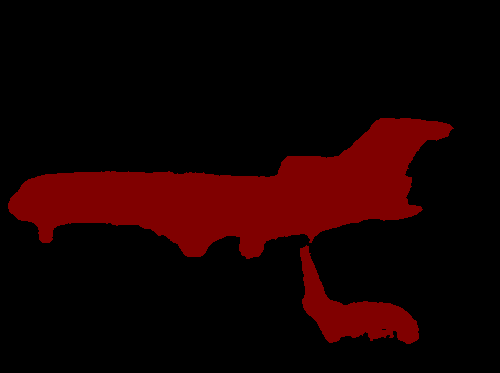}  \\ 
\includegraphics[width=0.16\linewidth] {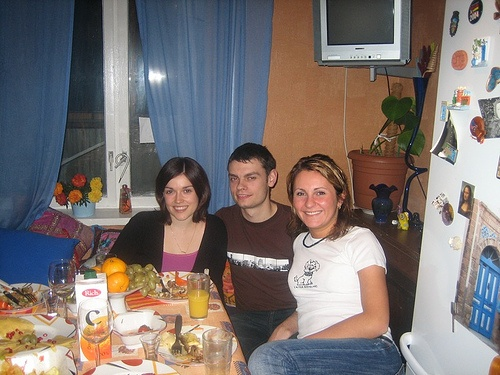}
\includegraphics[width=0.16\linewidth] {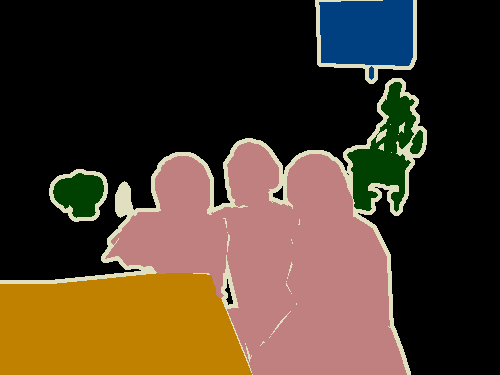}
\includegraphics[width=0.16\linewidth] {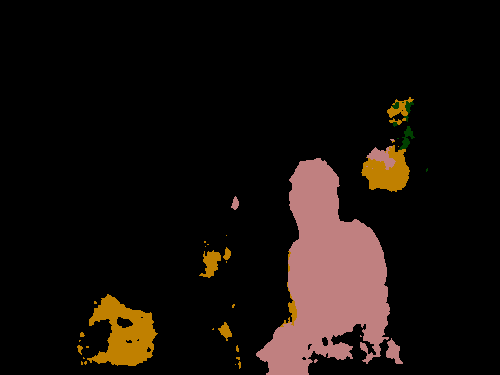}
\includegraphics[width=0.16\linewidth] {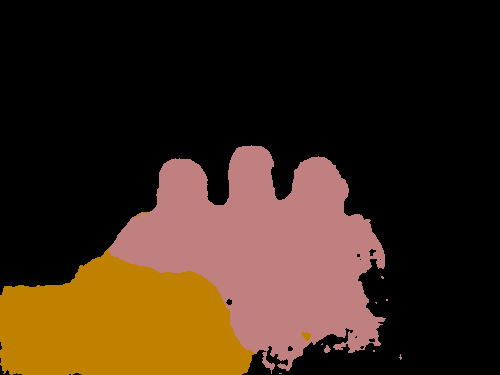}
\includegraphics[width=0.16\linewidth] {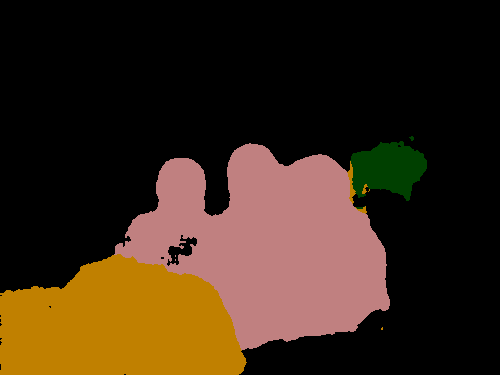}
\includegraphics[width=0.16\linewidth] {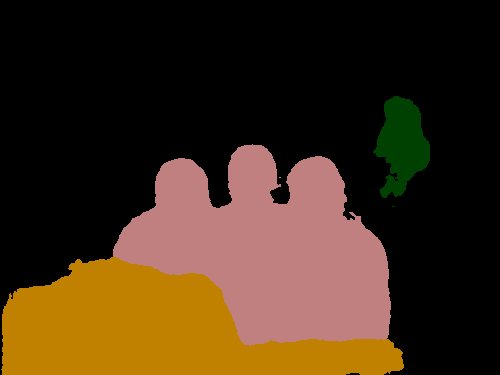}  
\end{minipage}
\caption{
Semantic segmentation results of several PASCAL VOC 2012 validation images based on the models trained on a different number of pixel-wise segmentation annotations. 
}
\label{fig:qualitative}
\end{figure}

Figure~\ref{fig:qualitative} presents several qualitative results from the proposed algorithm.
Note that our model trained only with five strong annotations per class already shows good generalization performance, and that more training examples with strong annotations improve segmentation accuracy and reduce label confusions substantially.
Refer to our project website\footnote{\url{http://cvlab.postech.ac.kr/research/decouplednet/}} for more comprehensive qualitative evaluation.

%%%%%%%%%%  Segmentation Results  - 6 Column   %%%%%%%%%%%%

%% file: conclusion.tex
% !TEX root = nips2015_transfer.tex
\section{Conclusion}
\label{sec:conclusion}

We proposed a novel deep neural network architecture for semi-supervised semantic segmentation with heterogeneous annotations, where classification and segmentation networks are decoupled for both training and inference. 
%By decoupling these two objectives, training and inference are performed independently by individual networks for classification and segmentation.
%By decoupling these two objectives, we can achieve semantic segmentation by learning individual networks for classification and segmentation separately;
%it simplifies overall training procedures, and conceptually appropriate to train the networks using heterogeneous and unbalanced training data with image-level and pixel-wise annotations.
The decoupled network is conceptually appropriate for exploiting heterogeneous and unbalanced training data with image-level class labels and/or pixel-wise segmentation annotations, and simplifies  training procedure dramatically by discarding complex iterative procedures for intermediate label inferences.
%Our algorithm works very well even with a handful number of training examples with segmentation annotations.
Bridging layers play a critical role to reduce output space of segmentation, and facilitate to learn segmentation network using a handful number of segmentation annotations.
Experimental results validate the effectiveness of our decoupled network, which outperforms existing semi- and weakly-supervised approaches with substantial margins.